\documentclass[runningheads]{llncs}
\usepackage[T1]{fontenc}
\usepackage{subcaption} 
\usepackage{tikz}
\usepackage{siunitx}
\usepackage{booktabs}
\usepackage{amsmath}
\usepackage{xcolor}     
\usepackage{hyperref}   
\usepackage{graphicx}
\definecolor{lime}{HTML}{A6CE39}
\DeclareRobustCommand{\orcidicon}{%
	\begin{tikzpicture}
	\draw[lime, fill=lime] (0,0) 
	circle [radius=0.16] 
	node[white] {{\fontfamily{qag}\selectfont \tiny ID}};
	\draw[white, fill=white] (-0.0625,0.095) 
	circle [radius=0.007];
	\end{tikzpicture}
	\hspace{-2mm}
}

\foreach \x in {A, ..., Z}{%
	\expandafter\xdef\csname orcid\x\endcsname{\noexpand\href{https://orcid.org/\csname orcidauthor\x\endcsname}{\noexpand\orcidicon}}
}

\begin{document}

\title{Predictive Maintenance for Ultrafiltration Membranes Using Explainable Similarity-Based Prognostics}
\titlerunning{Explainable RUL for Ultrafilteration}
%

\author{Qusai Khaled\inst{1}{\orcidA{}} \and
Laura Genga\inst{2}{\orcidC{}} \and
Uzay Kaymak\inst{1}{\orcidB{}}}

\authorrunning{Q. Khaled et al.}
%
\institute{%
Jheronimus Academy of Data Science, Eindhoven University of Technology, Eindhoven, The Netherlands\\
\email{qusai.khaled@ieee.org, U.Kaymak@ieee.org}
\and
School of Industrial Engineering, Eindhoven University of Technology\\
\email{l.genga@tue.nl}
}

\maketitle
\begin{abstract}

In reverse osmosis desalination, ultrafiltration (UF) membranes degrade due to fouling, leading to performance loss and costly downtime. Most plants rely on scheduled preventive maintenance, since existing predictive maintenance models, often based on opaque machine learning methods, lack interpretability and operator trust. This study proposes an explainable prognostic framework for UF membrane remaining useful life (RUL) estimation using fuzzy similarity reasoning. A physics-informed Health Index, derived from transmembrane pressure, flux, and resistance, captures degradation dynamics, which are then fuzzified via Gaussian membership functions. Using a similarity measure, the model identifies historical degradation trajectories resembling the current state and formulates RUL predictions as Takagi–Sugeno fuzzy rules. Each rule corresponds to a historical exemplar and contributes to a transparent, similarity-weighted RUL estimate. Tested on 12,528 operational cycles from an industrial-scale UF system, the framework achieved a mean absolute error of 4.50 cycles, while generating interpretable rule bases consistent with expert understanding. 

\end{abstract}

\section{Introduction}

In reverse osmosis (RO) desalination plants, ultrafiltration (UF) membranes serve as the critical first line of defense in pretreatment \cite{al2020ultrafiltration}, protecting downstream RO membranes from suspended solids, colloids, and particulate matter that accelerate membrane fouling and degradation. Over operational cycles, UF membranes undergo progressive deterioration characterized by reduced permeate flux and elevated transmembrane pressure (TMP), driven by irreversible pore blocking and fouling accumulation \cite{shi2014fouling}. While routine backwash cycles restore performance, membrane quality inevitably degrades beyond recovery, necessitating increasingly aggressive interventions—chemical cleaning, clean-in-place (CIP) protocols, and ultimately membrane replacement after hundreds of filtration cycles.

Current practice in water treatment and desalination relies on reactive or periodic maintenance strategies \cite{voipan2025hybrid}, where scheduled backwash and replacement intervals are predetermined based on conservative estimates or manufacturer guidelines \cite{oraby2025advances} \cite{molkeda2023corrective}. Although this approach mitigates catastrophic failure risk, it results in substantial inefficiencies: premature replacement of membranes with remaining service life, unanticipated failures between scheduled interventions, and suboptimal resource allocation \cite{tian2025predictive}. Predictive maintenance offers a paradigm shift by estimating the remaining useful life (RUL) of individual membrane modules based on real-time condition monitoring, enabling data-driven intervention scheduling that balances operational reliability with cost optimization \cite{alenezi2025artificial}.

The adoption of predictive maintenance in industrial desalination environments faces a critical barrier: the interpretability gap of conventional machine learning models \cite{gawde2024explainable}. While neural networks, support vector machines, and ensemble methods can achieve high predictive accuracy, their black-box nature obscures the reasoning process, making it difficult for plant operators to validate predictions against domain expertise or justify maintenance decisions to stakeholders. This lack of transparency undermines trust and limits practical deployment \cite{richards2023rewards}, particularly in safety-critical systems where unexplained recommendations may be disregarded.

Accordingly, this paper presents an explainable fuzzy similarity method for estimating remaining useful life of UF membranes. The methodology fuzzifies degradation indicators—normalized flux, TMP, and their temporal derivatives—into linguistically interpretable terms using Gaussian membership functions. RUL prediction is then performed through similarity-weighted aggregation with historical degradation trajectories. This formulation provides transparent, traceable predictions: operators can directly inspect which historical cases attribute to the current estimate and understand the degree of similarity between present and past degradation patterns. By bridging data-driven prognostics with human-interpretable reasoning, the proposed approach enhances both the technical rigor and practical applicability of predictive maintenance for UF membrane systems in desalination pretreatment. The remainder of this paper is organized as follows. Section 2 reviews UF membrane degradation and predictive maintenance. Section 3 describes the proposed explainable fuzzy similarity–based RUL framework. Section 4 presents the experimental results. Section 5 concludes the paper.

\section{Maintenance of Ultrafiltration Membranes}

\begin{figure}[!htbp]
    \centering
    \includegraphics[width=0.8\columnwidth]{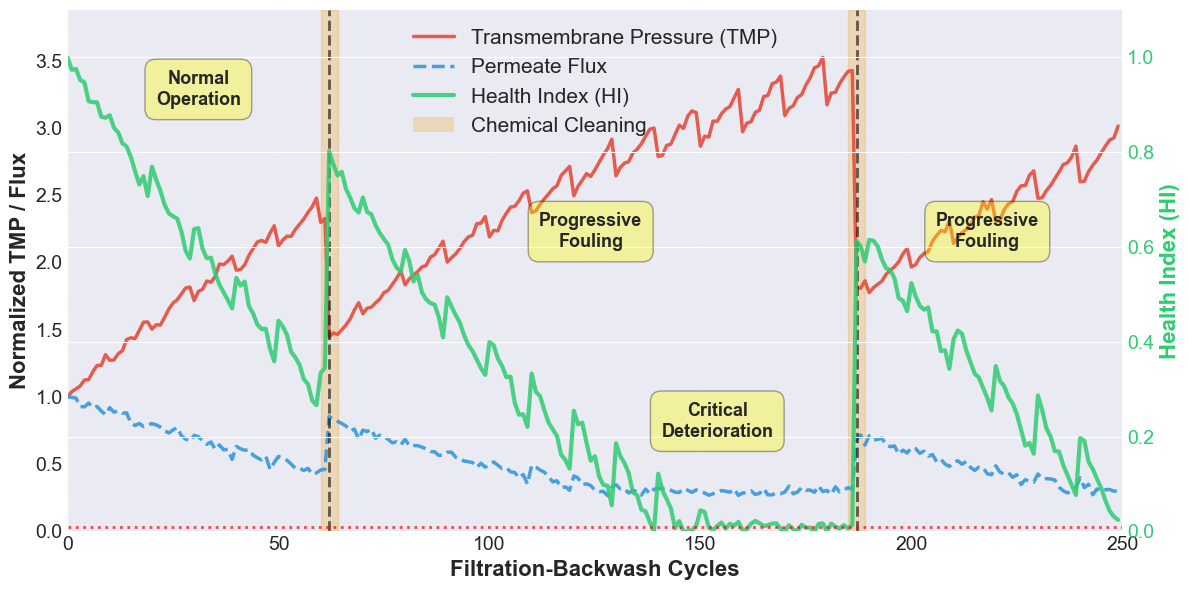}
    \caption{Conceptual degradation profile illustrating typical UF membrane behavior: progressive TMP rise, flux decline, and HI deterioration, with diminishing chemical cleaning effectiveness (orange) as irreversible fouling accumulates.}
    \label{fig:Degrade}
\end{figure}

UF membranes serve as critical pretreatment in reverse osmosis desalination plants, removing particulate matter and colloids that would otherwise foul downstream RO membranes, thereby extending their lifespan and reducing cleaning frequency \cite{lorain2007ultrafiltration}. However, they suffer progressive degradation from reversible and irreversible fouling mechanisms including cake-layer formation, pore blocking, and organic/inorganic adsorption. Fig.~\ref{fig:Degrade} illustrates this process through three indicators over filtration–backwash cycles: TMP, permeate flux, and Health Index (HI). Initially, TMP increases gradually while flux declines moderately—normal fouling recoverable through backwashing. With continued operation, irreversible fouling accumulates, causing accelerated TMP growth, sustained flux loss, and declining membrane health. Periodic chemical cleaning (orange dashed lines) temporarily restores performance but with diminishing effectiveness, reflecting permanent membrane deterioration. Eventually, the system enters critical degradation where cleaning fails to recover acceptable performance, necessitating membrane replacement.

\vspace{-0.5cm}
\vspace{-0.5cm}

Water treatment facilities traditionally rely on reactive or fixed preventive maintenance, causing unnecessary replacements, unexpected failures, and suboptimal resource allocation \cite{voipan2025hybrid}. Predictive maintenance enables proactive scheduling by analyzing real-time performance data, reducing unplanned downtime by 20–50\% and extending equipment lifespan \cite{alenezi2025artificial,ramzan2025literature}. Data-driven supervised models are increasingly applied to water-treatment assets for predicting failures like membrane fouling and pump wear \cite{alenezi2025artificial}. For example, \cite{zaveri2023towards} used KNN, random forest, ANNs and SVM to monitor fouling onset in solar desalination cells. In UF pretreatment, \cite{katibi2025insight} trained tree-based, ensemble, neural network and Gaussian-process models on 426-day UF data to forecast flow and fouling; their ensemble model enabled adaptive backwash scheduling, reducing TMP spikes and achieving $\approx$12\% cost savings. Similarity-based prognostic methods also show promise for RUL estimation due to strong generalization and simple updates \cite{xue2022improved}, as demonstrated by \cite{ince2020remaining} using random forests, gradient boosting and Gaussian processes. However, the black-box nature of high-performing ML models obscures their reasoning, hindering operator validation and maintenance decision justification in safety-critical systems.

Fuzzy systems provide explainable AI through interpretable linguistic variables and rule-based inference, bridging the gap between advanced ML models and human operators. Their ability to model imprecise data via IF-THEN rules has enabled diverse predictive maintenance applications \cite{el2025predictive}\cite{prommachan2022predictive}\cite{prommachan2024fuzzy}. \cite{zio2010data} pioneered fuzzy similarity-based RUL estimation for nuclear power plant components, using fuzzy distance measures to compare failure trajectories with reference patterns, achieving dynamic RUL updates through similarity-weighted aggregation. \cite{mayadevi2020predictive} developed a genetic fuzzy system for aircraft engine RUL estimation using NASA C-MAPSS turbofan data, learning interpretable rules via evolutionary optimization with competitive performance against state-of-the-art methods. In wind energy, \cite{zemali2023robust} integrated Kalman filters with ANFIS for fault detection across turbine subsystems, while \cite{garcia2025alarms} applied fuzzy logic to SCADA data for scalable online monitoring with reduced false alarms. These implementations leverage fuzzy membership functions to handle uncertainties and provide linguistic degradation interpretations, enabling operators to trace predictions to historical cases—crucial for safety-critical systems where black-box predictions undermine trust. However, fuzzy techniques remain underexplored in water treatment and UF systems, motivating this work.

\section{Explainable Ultrafiltration Prognostics}
\begin{figure}[!htbp]
    \centering
    \includegraphics[width=1.0\columnwidth]{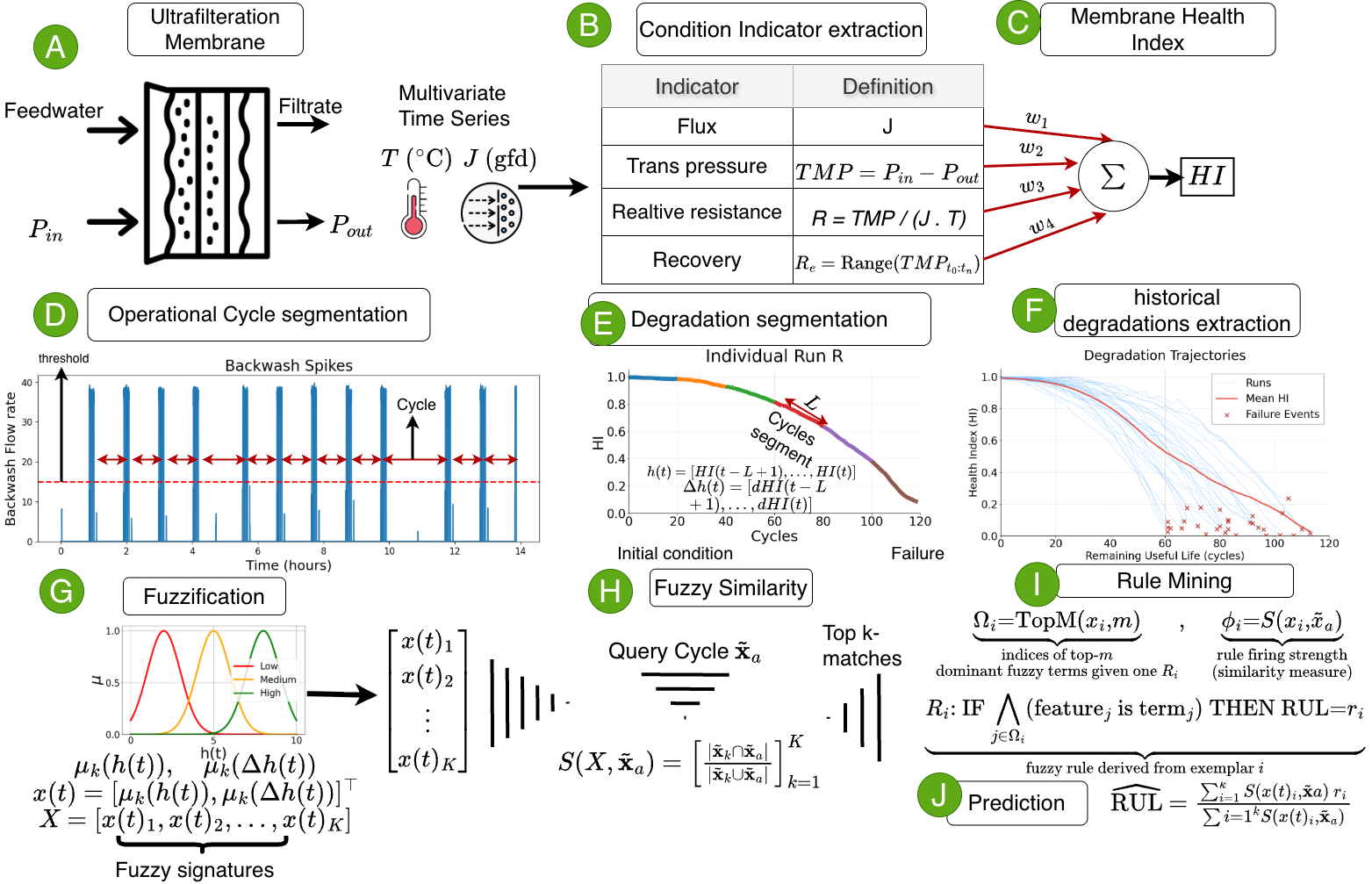}
    \caption{Explainable fuzzy similarity-based prognostic framework for UF membrane RUL estimation.}
    \label{fig:frama}
\end{figure}

The proposed explainable prognostic framework operates through a multi-stage pipeline that transforms raw sensor data into interpretable remaining useful life predictions. As illustrated in Fig~\ref{fig:frama}, the methodology begins with data acquisition from an operational ultrafiltration system (A), followed by physics-informed feature engineering to derive features relevant to membrane's health. These features are then aggregated as weighted sum into a composite Health Index (HI) (C) that quantifies membrane degradation state on a normalized scale from 0 (failed) to 1 (healthy). Operational cycles are then segmented based on backwash events (D) and further grouped into independent runs (E) representing distinct membrane lifecycles between replacements or chemical cleanings, from which historical failure trajectories are extracted (F). The core prognostic engine employs fuzzy logic to transform HI trajectories into 120-dimensional linguistic representations (Low, Medium, High) using Gaussian membership functions (G). For a given query cycle, the framework retrieves the top-k most similar historical degradation patterns from a pre-failure library using a set-theoretic similarity measure (H) \cite{setnes1998similarity}, formulating each match as a zero order Takagi–Sugeno fuzzy rule (I). The final prediction emerges through similarity-weighted aggregation (J), yielding both a RUL estimate and a fuzzy rule base with full traceability to similar runs. Each rule takes the form: IF [membrane state conditions] THEN RUL = X cycles (similarity = Y). Directly linking observable conditions to expected remaining life.

\subsection{Feature Engineering and Health Index Construction}

Membrane fouling is a multifaceted degradation process driven by hydraulic resistance buildup, flux decline, and feed water quality deterioration. To capture these interrelated phenomena, we derive a set of physics-informed features from raw sensor measurements, each reflecting a specific aspect of membrane performance and operating conditions. the derived features are transmembrane pressure (TMP), flux $J$, relative resistance $R$ and recovery $R_e$. When combined, these features can present a quantitative measure for the membrane health based on fouling conditions \cite{warsinger2017theoretical}. TMP represents The driving force for permeation and is quantified as the pressure differential across the membrane barrier as follows

{\footnotesize
\begin{equation}
\text{TMP} = P_{\text{feed}} - P_{\text{filtrate}},
\end{equation}
}

\noindent
where $P_\text{feed}$ measures feed-side pressure and $P_\text{filtrate}$ captures filtrate-side pressure. TMP inherently increases over operational time as fouling restricts flow pathways, making it a primary indicator of membrane condition. Next, flux $J$ is defined as normalized volumetric flux of permeate through the membrane, which is obtained directly from the filtrate flowmeter measurement. Flux decline under constant pressure operation signals progressive pore blockage and cake layer formation, representing the observable consequence of irreversible fouling accumulation \cite{shi2014fouling}. Then, Combining TMP and flux via Darcy's law for porous media yields an estimate of total membrane resistance

{\footnotesize
\begin{equation}
R_{\text{m}} = \frac{\text{TMP}}{J \cdot \mu_{\text{rel}} + \epsilon},
\end{equation}
}

\noindent
where $\mu_{\text{rel}}$ represents a linear correction factor defined as $\mu_{\text{rel}} = 1 - 0.02(T - 20)$, accounts for temperature-dependent viscosity variations, with $T$ in \textdegree C measured by sensors. This normalization ensures that changes in resistance reflect fouling rather than thermal effects, while $\epsilon$ prevents division by zero. In addition to the membrane intrinsic parameters, measuring recovery within each filtration cycle reflects the effectiveness of backwash cycles, which in return reflects an additional insight into fouling reversibility, defined as,

{\footnotesize
\begin{equation}
\text{Recovery} = \text{TMP}_{\text{max}} - \text{TMP}_{\text{min}}.
\end{equation}
}

High recovery indicates effective hydraulic cleaning, while diminishing recovery over successive cycles signals transition from reversible to irreversible fouling. Consequently, The Health Index synthesizes these indicators into a normalized metric (0 = failed, 1 = healthy)
\begin{equation}
\text{HI} = 0.30 \,(1 - R_m^*) + 0.25 \,(1 - \text{TMP}^*) + 0.30 \, J^* + 0.15 \, \text{Rec}^*
\end{equation}
where $R_m^*, \text{TMP}^*, J^*, \text{Rec}^*$ denote the normalized values of membrane resistance, transmembrane pressure, flux, and recovery, respectively. Min--max normalization is applied per operational run. Weights prioritize resistance (30\%) and flux (30\%) as primary fouling indicators, with secondary contributions from TMP (25\%) and recovery (15\%). A healthy membrane (HI $\approx$ 1) exhibits low resistance, minimal pressure elevation, high flux, and strong backwash recovery. As fouling accumulates, HI declines monotonically, reflecting simultaneous resistance growth, pressure escalation, flux loss, and diminished cleanability. While feed water quality parameters could enrich this assessment, the minimal hydraulic feature set enables deployment in facilities with limited sensor infrastructure.

\subsection{Operational Cycle Segmentation}

Backwash cycles are identified via spike detection in the backwash flow rate signal, which exhibits distinct short-duration increases during hydraulic cleaning relative to normal filtration. Detected cycles are grouped into independent operational runs, where each run represents a complete membrane degradation lifecycle from initial deployment or major recovery to failure.

A new run is initiated when: (i) the Health Index exhibits an abrupt increase exceeding 0.5, indicating chemical cleaning or (ii) a time gap exceeding 24 hours occurs between consecutive cycles, signaling operational shutdown. Each run terminates upon reaching the failure threshold (HI $\le 0.01$) or when the next recovery event begins. This segmentation ensures statistical independence between degradation trajectories used for similarity-based prognostics.

Within each run, RUL represents the number of filtration--backwash cycles remaining before the membrane reaches failure (HI $\leq 0.01$). Since membrane degradation progresses through distinct phases—from healthy operation through reversible fouling to irreversible deterioration—HI values are normalized per run to account for baseline differences between membrane installations. RUL is thus expressed in cycles rather than time, quantifying operational capacity until performance recovery actions become necessary.

\subsection{Fuzzy Membership Function Design}
\label{subsec:fuzzy_membership_design}

Assessing the historical trajectory is a perquisite for predicting remaining useful life. As such, for a membrane at a given cycle at position $t$ within a run, the historical trajectory of HI values and their rates of change are represented as

{\footnotesize
\begin{equation}
\mathbf{h}(t) = [HI(t-L+1), \ldots, HI(t)], \qquad
\Delta \mathbf{h}(t) = [dHI(t-L+1), \ldots, dHI(t)],
\end{equation}
}
\noindent
where $t$ denotes the current cycle position within the run, $L$ is the sliding window length capturing recent degradation history for the previous $L$ cycles, and $dHI$ denotes the first-order difference in $HI$ between consecutive cycles. This representation encodes both the current membrane state and the degradation trend leading up to cycle $t$, with the goal of the  assessing instantaneous condition and recent deterioration behavior. To construct a fuzzy representation of the encoded membrane state, each degradation feature $x_{j}(t)$ at cycle $t$, where $j \in \{\text{HI}, \text{dHI}\}$, is transformed into a fuzzy linguistic representation using three Gaussian membership functions,

{\footnotesize
\begin{equation}
\mu_{k}(x_j)
=
\exp\!\left(-\frac{(x_j-c_{k})^2}{2\sigma_j^2}\right),
\qquad k \in \{\text{Low}, \text{Med}, \text{High}\}.
\end{equation}
}

\noindent
Here, $c_{\text{Low}}$, $c_{\text{Med}}$, and $c_{\text{High}}$ denote the centers of the three membership functions, respectively, and
\begin{equation}
\sigma_j = 0.5 \cdot \frac{(c_{\text{High}} - c_{\text{Low}})}{2},
\end{equation}
controls the membership function bandwidth. The fuzzy state of feature $x_{j}$ is represented by the vector of membership degrees

{\footnotesize
\begin{equation}
\tilde{\mathbf{x}}_{j}(t) =
\big[
\mu_{\text{Low}}(x_{j}),
\mu_{\text{Medium}}(x_{j}),
\mu_{\text{High}}(x_{j})
\big].
\end{equation}
}

By concatenating the fuzzy representations across both features (HI and $dHI$) over all $L = 20$ cycles in the sliding window, the membrane state at cycle $t$ is encoded as a 120-dimensional fuzzy signature

{\footnotesize
\begin{equation}
\tilde{\mathbf{X}}(t) =
[
\tilde{\mathbf{x}}_{\text{HI}}(t-L+1),
\tilde{\mathbf{x}}_{\text{dHI}}(t-L+1),
\ldots,
\tilde{\mathbf{x}}_{\text{HI}}(t),
\tilde{\mathbf{x}}_{\text{dHI}}(t)
].
\end{equation}
}

This 120-dimensional vector ($20$ cycles $\times$ $2$ features $\times$ $3$ membership values) captures the recent degradation trajectory of the membrane, encoding both its instantaneous health and short-term deterioration trends in a form directly usable for fuzzy similarity comparison and rule-based reasoning.

\subsection{Fuzzy Similarity Estimation}
\label{subsec:fuzzy_similarity}

To quantify the resemblance between degradation trajectories, we compute the similarity between two fuzzy signatures, $\tilde{\mathbf{X}}_a$ and $\tilde{\mathbf{X}}_b$, using the Jaccard Index set-theoretic fuzzy similarity measure~\cite{setnes1998similarity}, given by

{\footnotesize
\begin{equation}
\label{eq:fuzzy_similarity}
S(\tilde{\mathbf{X}}_a, \tilde{\mathbf{X}}_b)
=
\frac{\left| \tilde{\mathbf{X}}_a \cap \tilde{\mathbf{X}}_b \right|}
     {\left| \tilde{\mathbf{X}}_a \cup \tilde{\mathbf{X}}_b \right|}
=
\frac{\sum_{k=1}^{120} \min\!\left(\mu_a^{(k)}, \mu_b^{(k)}\right)}
     {\sum_{k=1}^{120} \max\!\left(\mu_a^{(k)}, \mu_b^{(k)}\right) + \epsilon},
\end{equation}
}

where $\mu_a^{(k)}$ and $\mu_b^{(k)}$ denote the $k$-th membership degree in signatures $a$ and $b$, respectively. The numerator computes the component-wise minimum (fuzzy intersection), while the denominator computes the component-wise maximum (fuzzy union). The constant $\epsilon = 10^{-9}$ prevents division by zero. This measure ranges from 0 (completely dissimilar) to 1 (identical degradation patterns). High similarity values indicate that the current membrane degradation pattern closely resembles a given historical trajectory.

\subsection{RUL Estimation as a Takagi--Sugeno Fuzzy System}

A library of historical degradation patterns is constructed from training runs that reached failure (HI~$\leq$~0.01 sustained for at least one cycle). For each cycle in a given run, its fuzzy signature (a 120-dimensional vector encoding the memberships of HI and $dHI$ over the past $L=20$ cycles) and its corresponding singleton RUL are stored. Given a query cycle $q$, its fuzzy signature is compared against all library signatures $\{s_1, \ldots, s_m\}$ using \eqref{eq:fuzzy_similarity}, top-$k = 10$ most similar signatures are retrieved and each is treated as a Takagi--Sugeno rule, forming a compact, data-derived rule base:

{\footnotesize
\[
\text{Rule } i:~ \text{IF input resembles } s_i~ \text{THEN RUL = } r_i~(\text{cycles}),
\]
}
\noindent
where $r_i$ is the crisp RUL value associated with exemplar $s_i$. The predicted RUL is obtained through similarity-weighted TSK aggregation:

{\footnotesize
\begin{equation}
\label{eq:rul_estimation}
\widehat{\text{RUL}}
=
\frac{\sum_{i=1}^{k} S(q, s_i)\, r_i}
     {\sum_{i=1}^{k} S(q, s_i)},
\end{equation}
}
where $S(q,s_i)$ is the fuzzy similarity between the query and rule $i$.  
This yields a crisp, interpretable RUL estimate while grounding the inference process in a fuzzy rule base extracted directly from historical degradation behavior. Each of the top-$k$ retrieved historical cases functions as an active TSK fuzzy rule with a clear IF-THEN structure. The antecedent ("IF part") describes the degradation pattern, while the consequent ("THEN part") provides the observed RUL outcome from that historical case. For example,

{\footnotesize
\begin{quote}
\textbf{Rule 3:} IF \emph{recent health is declining} AND \emph{current health is moderate} THEN RUL = 24 cycles (similarity: 0.87).
\end{quote}
}

The similarity score $S(q, s_i)$ serves as the rule's firing strength, quantifying how closely the current membrane condition matches this historical pattern. Rules with higher similarity contribute more weight to the final prediction.

\subsubsection*{Rule Mining}

To make rule antecedents human-readable, it is recommended to maintain the size of rule base as small as possible, preferably between 5-7 antecedents \cite{khaled2025interpretable}. Thus we identify the most characteristic fuzzy memberships within each historical exemplar. Rather than presenting all 120 membership values, we extract the top 5 most dominant linguistic terms—those with the highest membership degrees—that best characterize that degradation pattern. For instance, if an exemplar's signature shows strong memberships in ``HI at cycle 20 is Medium'' ($\mu = 0.92$), ``HI at cycle 4 is High'' ($\mu = 0.88$), and ``dHI at cycle 19 is Low'' ($\mu = 0.85$), these become the rule's antecedent. This compression highlights the critical features defining each degradation mode without overwhelming the operator.

\subsection{Data and Experimental Design}
The deployed dataset~\cite{cohen2021uf} contains long-term operational data from an ultrafiltration (UF) system used as pretreatment for seawater reverse osmosis feedwater, collected during a multi-year field study at Port Hueneme, California, spanning 422 days of operation. The dataset captures both short- and long-duration filtration–backwash cycles under varying water quality conditions and coagulant dosing strategies. It comprises 24 process variables that describe the hydraulic, operational, and water quality characteristics of the UF system. Hydraulic and operational variables include UF inflow rate and element-specific inflow rates, microfilter inlet and trans-filter pressures, UF inlet and feed-side backwash pressures, UF filtrate-side pressure, and UF backwash flow rate. Water quality and chemical treatment conditions are represented through feedwater and filtrate turbidity, filtrate pH, filtrate temperature, coagulant dose, and chlorophyll concentration. Additional operational indicators, such as initial UF membrane resistance and feed pump RPM, support detailed analysis of membrane performance, fouling behavior, and backwash efficiency. For the experiments in this study, backwash cycles were detected from the backwash flow rate sensor using a fixed threshold of 15 gallons per minute, resulting in 12,501 valid operational cycles. These cycles were grouped into 373 independent operational runs.  Runs were then split chronologically into training (80\%) and test (20\%) sets, yielding 298 training runs (9,833 cycles) and 75 test runs (2,668 cycles). A sliding window of 20 cycles was applied to capture recent degradation history, and fuzzy membership functions were defined with Gaussian widths scaled to 50\% of the Low–High feature range. RUL values ranging from 0 to 93 cycles. To address class imbalance in the training data, fuzzy membership functions were configured using uniform grid partitioning rather than data-driven percentiles. This yielded HI centers at $[0.000, 0.500, 1.000]$ ($\sigma = 0.250$) and dHI centers at $[-1.000, 0.000, 1.000]$ ($\sigma = 0.500$), corresponding to equally spaced \textit{Low}, \textit{Medium}, and \textit{High} linguistic terms. This design choice prioritized discriminative power across the full operational range---from failed (HI $\approx 0$) to healthy (HI $\approx 1$) states---over fitting the skewed empirical distribution where 66\% of cycles exhibited HI $> 0.8$. The resulting 120-dimensional fuzzy signatures encoded both instantaneous membrane condition and temporal deterioration trends over 20-cycle sliding windows.

\section{Results and Discussion}
\label{sec:results}
\begin{figure}[h!]
\centering
\includegraphics[width=1.0\linewidth]{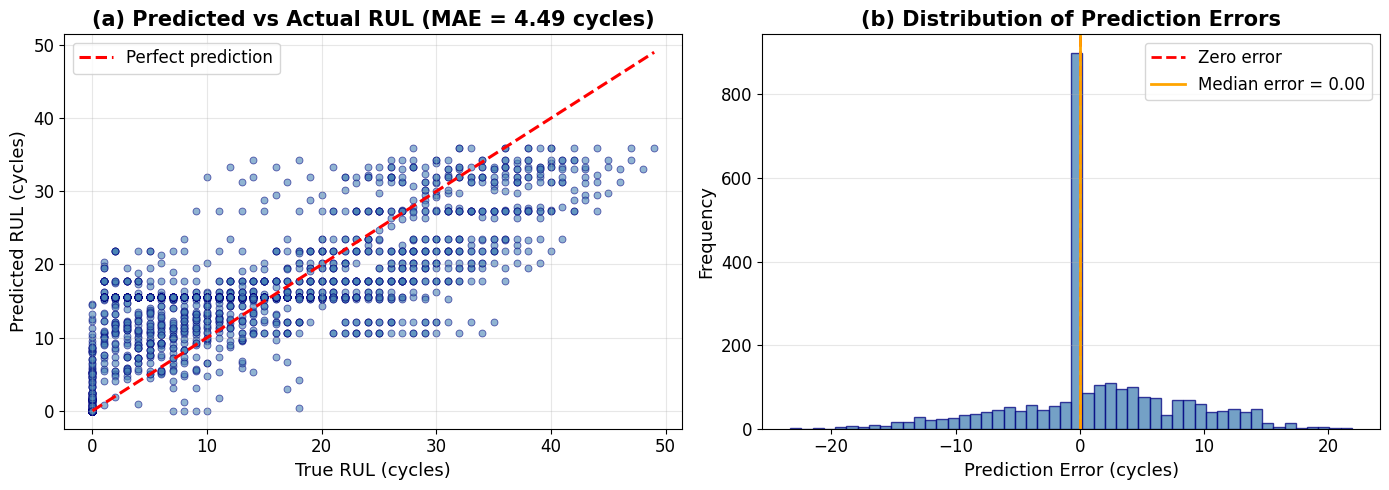}
\caption{(a)Predicted vs.\ actual RUL for 2,668 test cycles (MAE = 4.08 cycles). Predictions cluster around the ideal diagonal (red dashed line), with optimal accuracy in the medium-term range (6--15 cycles). Scatter increases for imminent failures (lower left) and long-term projections (upper right), reflecting trade-offs in uniform fuzzy partitioning design. (b)Distribution of prediction errors showing near-zero median error. The sharp central peak corresponds to accurate predictions across the dominant 6--15 cycle range, while distribution tails reflect increased uncertainty for extreme RUL values.}
\label{fig:pred_actual}
\end{figure}

The framework achieved a mean absolute error (MAE) of 4.08 cycles and a root mean square error (RMSE) of 6.28 cycles across the 2,668-cycle test set, with 80\% prediction-interval coverage of 68.6\%. Results shown in Figure \ref{fig:pred_actual} (a) present good predictive accuracy, with the majority of predictions clustering near the ideal diagonal. Figure \ref{fig:pred_actual} (b) presents the median prediction error of zero cycles, confirming unbiased estimation, though the distribution exhibits significant variance at both tails, reflecting inherent uncertainty in matching diverse degradation trajectories. Given a mean cycle duration of 121~s, this corresponds to an average temporal prediction error of approximately 8.2~minutes, enabling maintenance-scheduling accuracy comfortably within a single operational hour. The model produces RUL estimates up to 93 cycles in advance ($\approx 3.1$~hours of operational lookahead); however, accuracy is horizon-dependent (see Table~\ref{tab:stratified}), with strongest performance in the medium-term range (6--15 cycles, MAE $\approx 3.67$ cycles) and reduced precision for very short and very long horizons.

To place the quantitative results of our study into a comparative context, in industrial ultrafiltration practice, chemical cleaning interventions are typically triggered after a limited number of filtration--backwash cycles, with reported intervals on the order of tens of cycles depending on feedwater quality and operating conditions \cite{gu4793091evaluation,tang2022membrane}. In this context, the proposed model’s ability to forecast membrane end-of-life several dozen cycles ahead, with an overall MAE of approximately 4 cycles and even lower error in the operationally critical 6--15 cycle window, represents an improvement over conventional fixed-interval scheduling. Practically, this horizon-dependent performance can translate into actionable 20--30 cycle advance warnings ($\approx 40$--60~minutes) in many settings—sufficient time for chemical preparation and coordinated intervention.

\begin{table}[h!]
\centering
\caption{Stratified predictive performance by RUL horizon.}
\label{tab:stratified}
\scriptsize
\begin{tabular}{lccccc}
\toprule
\textbf{RUL Horizon} & \textbf{Cycles} & \textbf{MAE} & \textbf{RMSE} & \textbf{Coverage} \\
\textbf{(cycles)}   & \textbf{(n)}    & \textbf{(cycles)} & \textbf{(cycles)} & \textbf{(\%)} \\
\midrule
0--5   & 371  & 6.11 & 7.54  & 33.4 \\
6--15  & 611  & 3.67 & 4.77  & 64.8 \\
16--30 & 528  & 7.22 & 8.71  & 55.5 \\
30+    & 239  & 9.28 & 10.70 & 48.1 \\
\midrule
\textbf{Overall} & 2668 & 4.08 & 6.28 & 68.6 \\
\bottomrule
\end{tabular}
\end{table}

However, predictive accuracy varied substantially across RUL horizons, as shown in Table~\ref{tab:stratified}, revealing a critical trade-off inherent to the uniform fuzzy partitioning strategy. The framework excelled in the medium-term range (6--15 cycles: MAE = 3.67 cycles, coverage = 64.8\%), where operators benefit most from advance warning for maintenance scheduling. Notably, short-term prediction performance improved significantly compared to baseline percentile-based partitioning, achieving MAE = 6.11 cycles for imminent failures (RUL $\leq 5$), though coverage remained limited (33.4\%). Performance gains come at the expense of longer-term accuracy: predictions for RUL $> 15$ cycles exhibited degraded performance (MAE = 7.22--9.28 cycles), as the uniform partitioning reduced sensitivity to subtle differences in healthy membrane states.

\subsection{Interpretable Rule Generation}
\label{subsec:interpretable_rules}

\begin{figure}[h!]
\centering
\includegraphics[width=0.75\linewidth]{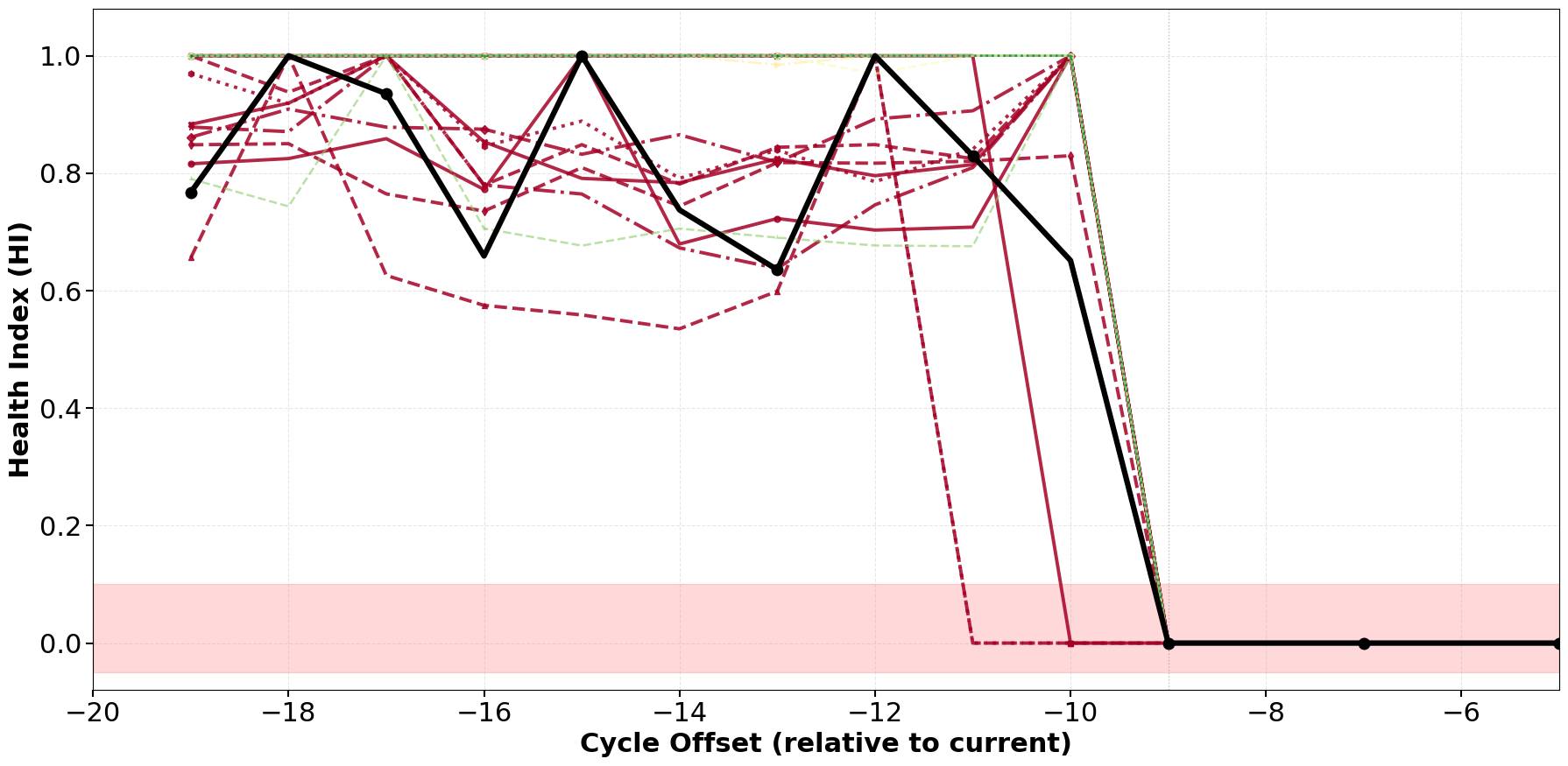}
\caption{Example degradation trajectory for query cycle 12528 showing Health Index of query cycle in black and top similar trajectories.}
\label{fig:query_window}
\end{figure}

An illustrative query cycle where ground truth RUL = 11 cycles is shown in Figure \ref{fig:query_window}. The plot depicts query cycle plot in black against other top 10 similar historical cases in the fuzzy signatures space shown in red. The HI trajectory enters a sustained near-zero regime after 10 cycles, indicating irreversible loss of membrane permeability and dominance of fouling/structural damage mechanisms. The top matches exhibit similar behavior and their mean values are relatively near their query. Aggregating the similar trajectories results in a predicted weighted RUL $\approx$ 12.7 cycles. From the derived trajectories, an example representative fuzzy rule contributing to this estimate is

\begin{quote}
\textbf{Rule (Similarity = 0.997):}  
IF $\mathrm{HI}_{t}$ is Very High (1.00) AND $\mathrm{dHI}_{t}$ is Medium (1.00) AND  
$\mathrm{HI}_{t-1}$ is Very High (0.90) AND $\mathrm{dHI}_{t-1}$ is Medium (0.90) AND  
$\mathrm{dHI}_{t-2}$ is Medium (0.81),  
THEN RUL = 11 cycles.
\end{quote}

This rule describes a membrane that remains in a healthy state over the last three cycles ($t, t\!-\!1, t\!-\!2$) while exhibiting a consistent medium-rate decline in HI, indicating gradual degradation. In the proposed framework, several such rules are activated in parallel and combined using similarity-weighted averaging, forming a mixture-of-experts structure in which each historical trajectory contributes partially to the final RUL estimate.


\section{Conclusion}

This study presented an explainable fuzzy similarity-based framework for ultrafiltration membrane RUL estimation, showing that interpretability and predictive accuracy can coexist in prognostic applications of water treatment. By fuzzifying physics-informed degradation indicators into linguistic representations and retrieving similar historical trajectories via set-theoretic fuzzy similarity measures, the framework achieved a mean absolute error of 4.08 cycles while providing Takagi–Sugeno rule bases traceable to specific operational precedents. the use of linguistic terms (Low, Medium, High) corresponds to how operators naturally discuss membrane health, avoiding abstract numerical outputs; operators can identify which recent cycles drove the prediction, supporting proactive intervention. Unlike black-box models that provide only a numerical RUL estimate, this approach reveals the underlying reasoning process while maintaining the predictive accuracy of data-driven methods. Deployment requires only four hydraulic sensors, enabling adoption in resource-constrained facilities. Future research would explore applications for reverse osmosis and microfilteration membranes.

\begin{credits}

\subsubsection{\discintname}
The authors have no competing interests to declare that are
relevant to the content of this article. 
\end{credits}
%
%
%
%

\bibliographystyle{splncs04}
\bibliography{ref}

\end{document}